\documentclass{llncs}

\usepackage[numbers,sort,sectionbib]{natbib}
\makeatletter
\renewcommand\@biblabel[1]{#1.}
\makeatother

\usepackage{amsmath,amssymb}
\usepackage[nocompress]{cite}
\usepackage{algorithm}
\usepackage{algorithmic}
\usepackage{caption}
\usepackage{ltablex}
\usepackage{booktabs}
\usepackage{hyperref}
\usepackage{subcaption}
\usepackage{url}
\usepackage{color}
\usepackage{graphicx}
\usepackage{graphicx}
\usepackage{array}
\usepackage{multicol}

\usepackage{makecell}

\usepackage[bottom]{footmisc}

\usepackage{array}
\newcolumntype{L}[1]{>{\raggedright\let\newline\\\arraybackslash\hspace{0pt}}m{#1}}
\newcolumntype{C}[1]{>{\centering\let\newline\\\arraybackslash\hspace{0pt}}m{#1}}
\newcolumntype{R}[1]{>{\raggedleft\let\newline\\\arraybackslash\hspace{0pt}}m{#1}}
\newcommand{\bx}{\ensuremath{\mathbf{x}}}

\newcommand{\ba}{\ensuremath{\mathbf{\alpha}}}
\newcommand{\bX}{\ensuremath{\mathbf{X}}}
\newcommand{\by}{\ensuremath{\mathbf{y}}}

\definecolor{dgreen}{RGB}{63, 175, 115}
\definecolor{orange}{RGB}{255,127,0}

\begin{document}

\title{Realistic risk-mitigating recommendations via inverse classification}
\author{Michael T. Lash
	\and
	W. Nick Street
}
\institute{University of Iowa, Iowa City, USA\\
	\email{\{michael-lash, nick-street\}@uiowa.edu}
}

%\title{Realistic risk-mitigating recommendations via inverse classification}
%\author{Anonymized 1
%	\and
%	Anonymized 2
%}
%\institute{Anonymized Institute\\
%	\email{\{Anonymized1, Anonymized2\}@anondomain}
%}

\maketitle

\begin{abstract}
Inverse classification, the process of making meaningful perturbations to a test point such that it is more likely to have a desired classification, has previously been addressed using data from a single static point in time. Such an approach yields inflated probability estimates, stemming from an implicitly made assumption that recommendations are implemented instantaneously. We propose using longitudinal data to alleviate such issues in two ways. First, we use past outcome probabilities as features in the present. Use of such past probabilities ties historical behavior to the present, allowing for more information to be taken into account when making initial probability estimates and subsequently performing inverse classification. Secondly, following inverse classification application, optimized instances' unchangeable features (e.g.,~age) are updated using values from the next longitudinal time period. Optimized test instance probabilities are then reassessed. Updating the unchangeable features in this manner reflects the notion that improvements in outcome likelihood, which result from following the inverse classification recommendations, do not materialize instantaneously. As our experiments demonstrate, more realistic estimates of probability can be obtained by factoring in such considerations.
\end{abstract}

\section{Introduction}

Individual-centric data mining and machine learning paradigms have the potential to enrich the lives of millions of people by first making accurate predictions as to their future outcomes and then secondly recommending implementable courses of action that make a desired future outcome more likely. Consider semi-annual medical physicals that assess the current status of an individual's well-being, for example, Patient 8584, taken from our experiments. We have information on three semi-annual medical examinations for this patient and know that, sometime between the second and third visit, this person was diagnosed with cardiovascular disease (CVD). After learning a predictive model and evaluating the patient at the first visit, their predicted probability of CVD was 22\%. Their predicted probability at the second visit was 36\%. Merely observing this individual's progression towards this negative life-altering event is informative, but unhelpful: crucial steps detailing how such risk can be reduced are needed.

In this work we propose using inverse classification, the process of making recommendations using a machine learning method to optimally minimize or maximize the probability of some outcome, in conjunction with longitudinal data. Specifically, we want to minimize patient 8584's probability of CVD beginning from the first medical exam we have on record. This is the first contribution of this work: to methodologically incorporate longitudinal data into the inverse classification process. 

Considerations we make, leading to more realistic assessments of risk and more meaningful recommendations, constitute our second and third contributions. Specifically, as demonstrated by Patient 8584, past medical visits have a bearing on feature patient visits and, ultimately, on whether or not there is an unfavorable outcome; cumulative actions made over a period of time led to this person experiencing an adverse event. Therefore, we propose to incorporate the predicted risk at previous visits as a feature in future visits. Doing so makes estimates of current risk more realistic and allows the inverse classification process to make recommendations that account for past behaviors.

Furthermore, when making recommendations using inverse classification, and subsequently estimating the probability of a particular outcome, we can further obtain a more realistic estimate of risk by using what are referred to as the immutable feature-values observed at the proceeding medical visit. Immutable features are attributes that can't be changed, such as one's age. We include these values from future visits in the estimation of risk to make the assessment more realistic. The inclusion of such future feature values reflects the fact that changes made by an individual, such as Patient 8584, are not made instantaneously (contribution three), but take time to implement.

The rest of the paper proceeds by first discussing related work in Section 2, our proposition on the incorporation of longitudinal data into the inverse classification process, and how we, specifically, make our three contributions, in Section 3. In Section 4 we discuss our experiment process, parameters, and results, demonstrating these three contributions (and a fourth tertiary contribution, discussed later). Section 5 concludes the paper.

\section{Related Work}

Emergent data mining and machine learning research involving \textbf{longitudinal data} is focused on methodologically leveraging such data, as well as the specific domains in which such methods can be employed. In \citep{Razavian2016} the authors explore deep neural networks that, with minimal preprocessing, learn a mapping from patients' lab tests to over 130 diseases (multi-task learning). In \citep{Wang2014} unsupervised learning methods are applied to longitudinal health data to learn a disease progression model. The model can subsequently be used to aid patients in making long-term treatment decisions. These works exemplify the way in which models can be learned to aid in predicting disease \citep{Razavian2016} and in forecasting disease progression \citep{Wang2014}. In this work we examine how (1) coupling longitudinal data with predictive models can make disease risk estimation more accurate and (2) how predictive models that incorporate historical risk and past behavior can be used to make recommendations that optimally minimize the likelihood of developing a certain disease.

\textbf{Inverse classification} methods are varied in their approach to finding optimal recommendations, either adopting a greedy \citep{Aggarwal2010,Chi2012,Mannino2000,Yang2012} or non-greedy formulation \citep{Barbella2009,Pendharkar2002,Lash2016,Lash2016b}. Past works also vary in their implementation of constraints that lead to more realistic recommendations, either being completely unconstrained \citep{Aggarwal2010,Chi2012,Yang2012}, or constrained \citep{Barbella2009,Pendharkar2002,Mannino2000,Lash2016,Lash2016b}. In this work, we adopt the formulation and framework related by \citep{Lash2016,Lash2016b} which accounts for (a) the features that can and cannot be changed (e.g.~age cannot be changed, but exercise levels can), (b) varying degrees of change difficulty (feature-specific costs) and (c) a restriction on the cumulative amount of change (budget). As in \citep{Lash2016}, we implement a method that avoids making greedy recommendations while still accounting for (a), (b) and (c).

\section{Inverse Classification}

In this section we begin by discussion some preliminary notation, followed by our inverse classification framework, which we subsequently augment to account for past risk and missing features. 

\subsection{Preliminaries}

Let $\bX_v=[\bx_{v_1},\dots,\bx_{v_n}]^{\top} \in \mathbb{R}^{n \times p}$ denote a matrix of training instances and $\by_{v+1}=[y_{v_i},\dots,y_{v_n}]^{\top} \in \{0,1\}^{n}$ their corresponding labels at visit $v$, where $v=1,\dots,V$. Here, individual visits $v$ can be viewed as discrete, no-overlapping time units, where $\bX_v$ is observed at a discrete time unit and $\bX_{v+1}$ is observed at the next discrete time unit $v+1$. We note here that $X_v$ is observed at one visit and the event of interest $\by_{v+1}$ is observed at the next discrete time unit, namely $v+1$. We do this to reinforce the fact that we are interested in modeling how the current state of instances $\bX_v$ results in some not-to-distant future outcome. Additionally, let $\bX_{v+1} \subseteq \bX_v$ hold. That is, the instances at $v+1$ or also present at $v$, reflecting that these datasets are longitudinal.

Let $f_v: \mathbb{R}^p \rightarrow \mathbb{R}$ denote a trained classifier that has learned the mapping $\bX_v \rightarrow \by_{v+1}$. Once trained, $f_v(\bx)$ can be used to make a prediction as to the short-term outcome of test instance $\bx$ at visit $v+1$. Here, $f_v$ can be any number of classification functions.

With these preliminaries in mind we ultimately want to leverage longitudinal data to accomplish two things: \textbf{(1)} to obtain realistic recommendations for $\bx_{v}$ and \textbf{(2)} to use estimated risks from previous visits as predictive features in the present to further improve risk estimates of $\bx_{v}$.

%\begin{figure}
%	\centering
%	\includegraphics[]{graphics/new_long_data_risk_est_print.pdf}
%	\caption{Using predicted risk at a previous visit as a feature in future visits. \label{fig:risk_feature}}
%\end{figure}
%
%\subsection{Risk Estimation}
%
%
%To address (1), we propose using $r^i_v = f_v(x^i_v)$ and then incorporating $r_i$ as a feature of $\bx_i^{v+1}$ such that $\bX_{v+1}=[(r^1_{v},\bx^1_{v+1}),\dots,(r^n_{v},\bx^n_{v+1})]^{\top}$. Moreover, we also propose incorporating $r^i_v$ as a feature into all subsequent visits. This proposition can be observed graphically in Figure \ref{fig:risk_feature}.

\subsection{Inverse classification}

To address \textbf{(1)}, an optimization problem must be formulated.  Consider

\begin{align}
\min_{\bx} \hspace{1mm}& f_v(\bx_v) \\\nonumber
\text{s.t.} \hspace{1mm}& \textbf{c}^{\top}_v \mathbb{I} (\bx_v - \bar{\bx}_v) \leq B_v\\\nonumber
\hspace{1mm} & l_j \leq x_j \leq u_j, \text{ for } j=1,\dots,p
\label{eq:ic_1}
\end{align}

where $\bx$ is being optimized, $\bar{\bx}$ is the original instance, $\textbf{c}$ is a cost vector, $\mathbb{I}$ is a signature matrix where $\mathbb{I}_{jj} \in \{-1,1\}$, $B$ is a specified budget and $l_j,u_j$ are imposed lower and upper bounds, respectively. To control the direction of feature optimization we set 

\begin{align}
l_i/u_i = \left\{\begin{matrix}
\bx_{D_j}& \textrm{if } \mathbb{I}_{jj} = 1 / \mathbb{I}_{jj} = -1 \\ 
\min(0,\bx_{D_j}) / \max(1, \bx_{D_j}) & \textrm{otherwise}
\end{matrix}\right.
\end{align}

which has the effect of increasing (decreasing) features when $\mathbb{I}_{jj} = 1 \textrm{ (-1)}$.

In a realistic setting, such as the medical domain of our experiments below, not all $\bx_j$ are mutable. To reflect this, we partition the features into to two non-overlapping sets, $U$ and $C$, representing the immutable and mutable features, respectively. Not all mutable features can be changed directly, however. Some such features are altered indirectly.  To reflect this, we further partition $C$ into $D$ and $I$, reflecting those that are directly mutable and those that are indirectly mutable, respectively. Additionally, we define a function $\phi:\mathbb{R}^{|U|+|D|} \rightarrow \mathbb{R}^{|I|}$ that estimates the indirectly mutable features values using the features $U$ and $D$.  Incorporating these distinctions into \eqref{eq:ic_1} gives us 

\begin{align}
\min_{\bx_C} \hspace{1mm}& f_v(\bx_{v_U},\phi(\bx_{v_U},\bx_{v_D}),\bx_{v_D}) \\\nonumber
\text{s.t.} \hspace{1mm}& \textbf{c}^{\top}_v \mathbb{I} (\bx_v - \bar{\bx}_v) \leq B_v\\\nonumber
\hspace{1mm} & l_{v_j} \leq x_{v_j} \leq u_{v_j}, \text{ for } j \in D
\end{align}

where we are optimizing only the changeable features $D$.

\subsubsection{Past Risk}

\begin{figure}
	\centering
	\includegraphics[scale=1.15]{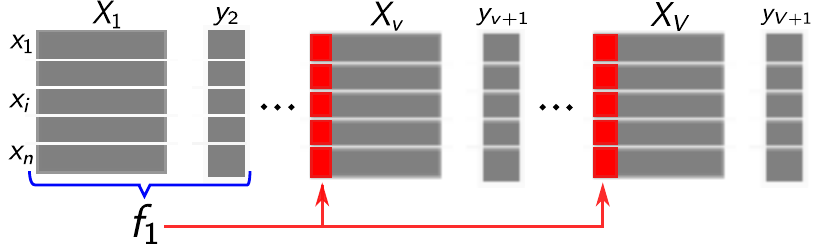}
	\caption{Using predicted risk at a previous visit as a feature in future visits. \label{fig:risk_feature}}
\end{figure}

To address \textbf{(2)}, we propose using $r_{v-1} = f_{v-1}(\bx_{v-1})$ and then incorporating $r_{v-1}$ as a feature of $\bx_{v}$ such that $\bx_{v}=[(r_{1},\cdots,r_{v-1},\bx_{v_U}),\bx_{v_I},\bx_{v_D}]^{\top}$. We believe that the inclusion of past risk as features in the present will lead to more accurate probability estimates and, as a result, better inform the inverse classification process. Our proposition to include past risk as predictive features in the present can be observed graphically in Figure \ref{fig:risk_feature}, albeit in simplified form.

\subsubsection{Missing Feature Estimators}

Practically speaking, features $S$ at one $v$ are not always present in a subsequent time period. Namely, $S_{v=k} \neq S_{v=q}$ where $k,q \in \{1,\dots,V\}$. To overcome this issue, we propose defining the full set of features at $v=1$, such that $S_{v=q} \subseteq S_{v=1}$ and $S_{v=q} \cap S_{v=1}= S_{v=q}$. During subsequent visits $q$, we propose to use an estimator, using the subset of features available at $q$, but learned using $v=1$ data, to impute these values. To reflect this let $\psi_m:\mathbb{R}^{p_{v=q}} \rightarrow \mathbb{R}, \text{ for }m \in M_{v=q}$, where $\psi_m$ is either a regressor or a classifier, depending on the feature being estimated. This results in $S_{v=1} \equiv \{S_{v=q},S_{M_{v=q}}\}$

Taking past risk and the estimation of missing features into account, the optimization problem can be reformulated to

\begin{align}
\label{eq:final_opt}
\min_{[\bx_C,\bx_{C_M}]} \hspace{1mm}& f_v([\bx_U,\bx_{U_M}],\phi([\bx_U,\bx_{U_M}],[\bx_D,\bx_{D_M}]),[\bx_D,\bx_{D_M}]) \\\nonumber
\text{s.t.} \hspace{1mm}& c^{\top}_v \mathbb{I} ([\bx,\bx_M] - [\bar{\bx},\bar{\bx}_M]) \leq B_v\\\nonumber
\hspace{1mm} & l_j \leq x_j \leq u_j, \text{ for } j \in D
\end{align}

where we abuse the notation by letting $U_M$ and $D_M$ denote the missing features for each of the respective $U,D$ feature sets and assume that each of these missing feature sets proceeds those that are known. Also note that past risk features have been merged with $\bx_{v_U}$.

\subsubsection{Implementing Recommendations and Risk Estimates} 

\begin{figure}
	\centering
	\includegraphics[scale=1.05]{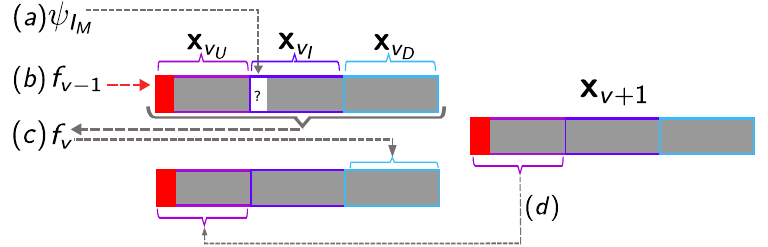}
	\caption{The inverse classification process using longitudinal instances. (a) Missing features are estimated (in this example there is a missing indirectly mutable feature). (b) Past risk is used as an immutable feature in the present. (c) Inverse classification is performed. (d) The immutable features at $v+1$ are used to estimate the new level of risk for the optimized test instance.\label{fig:ic_process}}
\end{figure}

Performing Equation \eqref{eq:final_opt} produces $\bx^*_{v}$, an optimized version of $\bx_v$ whose risk has been minimized. In the real world, however, it takes time to implement the feature updates that actually take $\bx_v$ to $\bx^*_{v}$. To reflect this, we replace $\bx_{v_U}^{*}$ with $\bx_{v+1_U}$ and update $\bx_{v_I}^{*} = \phi(\bx_{v+1_U},\bx_{v_D}^{*})$. In other words, we use the immutable features at the next discrete time unit, and a new estimate of the indirectly mutable, based on these updated immutable features, along with the optimized mutable features, to obtain a more realistic $\bx^*_{v}$. We fully outline the inverse classification process in Figure \ref{fig:ic_process}.

Finally, performing inverse classification in the manner we have outlined requires an optimization mechanism, of which there are many. To further zero in on a suitable method, we outline an assumption and the desirable qualities we'd like the optimization method to have. First, as in \citep{Lash2016}, we assume that $f_v$ is differentiable and it's gradient is $L$-Lipschitz continuous. That is, $\|\nabla f_v(\bx)-\nabla f_v(\bx')\|\leq L\|\bx-\bx'\| \forall \bx,\bx'\in\mathbb{R}^{|D|}$. If we assume that $f_v$ is linear, and observing that our constraints in Equation \eqref{eq:final_opt} are also linear, the optimization can be performed efficiently.  However, we do not wish to make this prohibitive assumption, although we do wish to solve the problem efficiently. Therefore, we elect to use the projected gradient method \citep{Nesterov07composite,Ghadimi:13a}, which has been shown to have a convergence rate of $O(\frac{1}{t})$ for nonlinear problems whose constraints are linear.

\section{Risk Mitigating Recommendations}

In this section we outline the data used in our experiments, our experiment process, and results.

\subsection{Data}

Our data is derived from the Atherosclerosis Risk in Communities (ARIC) study \citep{aric1989}. The investigation began in 1987 wherein ~16000 individuals from four different communities were initially examined by physicians. Individual assessments entailed the thorough documentation of a variety of patient-characterizing attributes. These components can be categorized as demographic (e.g.,~age), routine (e.g., weight), lab-based (e.g.,~blood glucose levels), and lifestyle-based (e.g., amount of exercise). Subsequent follow-up exams were conducted on a bi-annual basis thereafter.
\begin{sloppypar}
ARIC data is freely available, but does require explicit permission-to-use prior to acquisition. Additionally, some processing and cleansing must be done. For the sake of reproducibility, the code that takes these data from their acquisition state to the final state used in our experiments is provided at {\sffamily \textcolor{dgreen}{github.com/michael-lash/LongARIC}}
\end{sloppypar}.

We use three sets of data in our experiments, representing three bi-annual physician visits. Our target variable is defined based on outcomes observed at $v+1$. These outcomes pertain to cardiovascular disease (CVD) events. We define a CVD event to be one of the following: probable myocardial infarction (MI), definite MI, suspect MI, definite fatal coronary heart disease (CHD), possible fatal CHD, or stroke. Patients having such an event recorded at their $v+1$ examination have $y_{v+1}=1$, whereas patients not having any such events at $v+1$ have $y_{v+1}=0$. Patients for whom one of these events are observed at a previous visit are excluded from subsequent visit datasets. We do this to ensure that we are continuing, at each visit, to learn a representation that is consistent with two-year risk of CVD. Therefore, $\bX_3 \subseteq \bX_2 \subseteq \bX_1$. Table \ref{tab:dset_sum} summarizes the number of patients, features/missing features, and positive instances at each of the visits.

\begin{table}[h]
	\centering
	\begin{tabular}{|l|c|c|c|}
		\hline
		Dataset & Instances & Feats/Missing & $y_{v+1}=1$ \\
		\hline
		$X_1$ & 12223 & 122/0 & 232 \\
		\hline
		$X_2$ & 11057 & 98/24 & 249 \\
		\hline
		$X_3$ & 9883 & 74/48 & 231 \\
		\hline	
	\end{tabular}
	\caption{ Dataset descriptors.\label{tab:dset_sum}}
\end{table}

\subsection{Experiment Information}

\subsubsection{Evaluative Process}

The process of conducting our experiments is carefully crafted 
such that no data used in making recommendations is ultimately used 
in evaluating their success (final probability estimates). As such, each of
the $X_v$ datasets are partitioned into two parts at random. One part is used 
to train an $f_v$ that is used for the inverse classification process. The 
second partition is further partitioned into 10 parts. One part is used as a 
test set for which recommendations are obtained, while the other nine are used to train a validation model that evaluates the probability of CVD resulting from the recommendations obtained. By maintaining 
partition separation, and iteratively cycling between the role each partition plays, 
recommendations can be obtained for all $X_v$ instances. The process is more definitively outlined in \citep{Lash2016}.

\subsubsection{Experiment Parameters}

In our experiments we choose to use RBF SVMs \citep{Boser1992} as our $f_v$, which can be trained by solving the dual optimization problem

\begin{align}
\label{SVMdual}
\max_{\ba\in\mathbb{R}^n}&\sum_{i=1}^n\alpha_i-\frac{1}{2}\sum_{i=1}^n\sum_{j=1}^n\alpha_i\alpha_jy^iy^jk(\bx^i,\bx^j)\\\nonumber
\text{s.t.}&\sum_{i=1}^n\alpha_iy^i=0\text{ and }0\leq \alpha_i\leq C\text{ for }i=1,2\dots,n
\end{align}

where $k(\bx,\bx')=\exp\left(-\frac{\|\bx-\bx'\|^2}{2\sigma^2}\right)$ is the Gaussian kernel, $\sigma>0$ and $\|\cdot\|$ represents the Euclidean norm in $\mathbb{R}^p$. Practically speaking, any number of other kernels could have been selected. We elect to use these as they were observed to have good inverse classification performance in \citep{Lash2016}. Furthermore, by employing Platt Scaling \citep{platt1999} we can directly learn a probability space that is more easily interpreted.

Secondly, we elect to use kernel regression \citep{Nadaraya1964,Watson1964} as our indirect feature estimator $\phi_v$, which is given by 

\begin{eqnarray}
\label{kernelreg}
\bx_{v_I} &=& \frac{\sum_{i=1}^{n} k([\bx^i_{v_D}, \bx^i_{v_U}],[\bx_{v_D}, \bx_{v_U}])\bx_{v_I}^i}{\sum_{i=1}^{n} k([\bx^i_{v_D}, \bx^i_{v_U}],[\bx_{v_D}, \bx_{v_U}])},
\end{eqnarray}

where the $k(\bx,\bx')=\exp\left(-\frac{\|\bx-\bx'\|^2}{2\sigma^2}\right)$ Gaussian kernel and the value $\sigma > 0$ is selected based on cross-validation. We elect to use this method as the estimation is based on point similarity, which is consistent with our assumption that similar points will have similar CVD probabilities.

\subsubsection{Experiments}

Using this evaluative procedure and outlined learning methods, we propose three experiments. Two of these experiments directly address how the inclusion of past risk and $\bx_{v+1_U}$ affect probability estimation, while the third is somewhat tertiary to this objective.

\textbf{Experiment 1: }We demonstrate that more realistic estimates of probability, following the application of the inverse classification process, can be obtained by leveraging longitudinal data. We do so by first performing inverse classification on each of the three $X_v$ and estimate the resulting probabilities independent of one another. Then, we do the same procedure, except we use $\bx_{v+1_U}$ instead of $\bx_{v_U}$, and $\bx_{v_I}= \phi(\bx_{v+1_U},\bx^{*}_{v_C})$ instead of $\bx_{v_I}= \phi(\bx_{v_U},\bx^{*}_{v_C})$ to estimate the probabilities. This is related by (d) in Figure \ref{fig:risk_feature}.

\textbf{Experiment 2: }We examine the collective impact of including past risk as a predictive feature in the present, related by (b) in Figure \ref{fig:risk_feature}, in conjunction with $\bx_{v+1_U}$ in performing inverse classification and assessing the resulting outcome probabilities.

\begin{figure}[H]
	\centering
	\includegraphics[scale=.90]{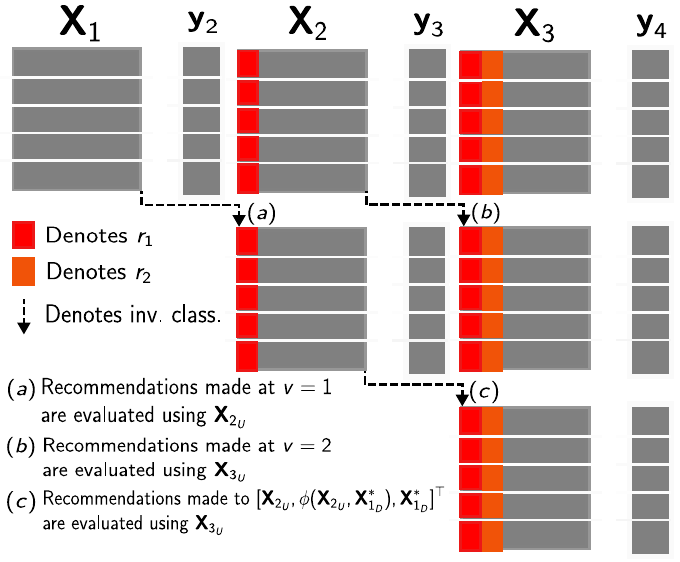}
	\caption{The inverse classification process for longitudinal instances. \label{fig:experiments}}
\end{figure}

Experiment 2 is outlined by Figure \ref{fig:experiments}.

\textbf{Experiment 3: }We would like to see whether the use of a learning method is better suited to estimating missing features in subsequent visits, or whether a simple carry-forward procedure is better. In other words, can we simply use the known feature values at a previous visit $v-1$ to estimate missing features at $v$, or is it better to use a learning method? To test what method works best we randomly selected two continuous features and one binary feature that are known in $v=2$.  We then trained various classifiers/regressors (where appropriate), using the known $v=2$ features, but with $v=1$ data, along with the carry-forward procedure, to make estimates. We then observed what performed the best in terms of MSE (continuous) or AUC (binary).

\subsection{Results}

In this section we present the results of our three experiments, albeit somewhat out of order. We begin by showing the results of Experiment 3, wherein we discovered the best method of estimating missing features at future visits. We present this first, as the results are used in practice in the subsequent two experiments.

\begin{table}[h]
	\centering
	\begin{tabular}{|l|c|c|c|}
		\hline
		 \diaghead{\theadfont Diag CHeadssss  }{Method}{Feat/type}& Alcohol/cont & Statin Use/bin & Hematocrit/cont \\
		\hline
		Carry & 50.55 & .579  & 7.55  \\
		\hline
		RBF SVM & 47.65 & .50  & 9.31  \\
		\hline
		Lin SVM & 47.65 & .50 & 9.31 \\
		\hline
		CART & 30.97 & .569  & 2.37 \\
		 \hline	
		kNN & \textbf{29.37} & .50  & 12.81 \\
		  \hline
		Log Reg & NA & \textbf{.984}  & NA \\
		   \hline
		Ridge & \textbf{29.42}* & NA & \textbf{1.24} \\
		    \hline
	\end{tabular}
	\caption{ Missing feature estimation (Experiment 3) results given in MSE for continuous features and AUC for binary.\label{tab:miss_feat}}
\end{table}

Table \ref{tab:miss_feat} shows the results of applying the described carry-forward procedure and a host of learning algorithms to the three randomly selected features. In all cases at least one learning method outperforms the carry-forward procedure. For continuous features, we observe that Ridge regression works well, and that $k$NN is either the best or the worst model. Therefore, we estimate continuous features using Ridge regression. The binary feature is best estimated by logistic regression. We use these selected learning methods to estimate missing features at $v=2,3$.

\begin{figure}[H]
	\centering
	%\begin{subfigure}[]{.4\linewidth}
	\includegraphics[scale=.39]{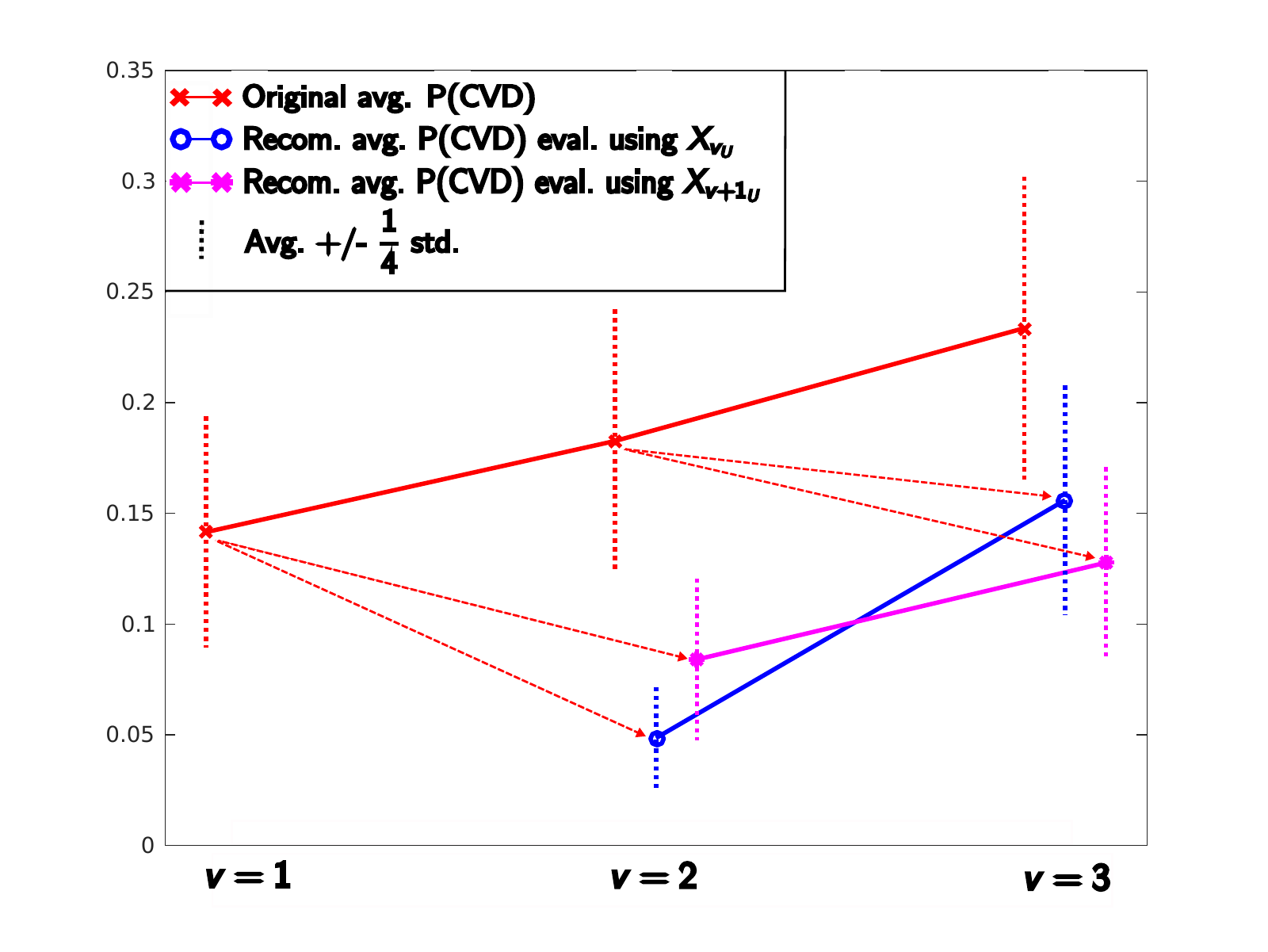}
	\caption{Results from Experiment 1. \label{fig:res_exp1}}
	%\end{subfigure}
	%\begin{subfigure}[]{.4\linewidth}
	%	\includegraphics[scale=.30]{figures/risk_opt_plot_new.pdf}
	%	\caption{Results from Experiment 2. \label{fig:res_exp2}}
	%\end{subfigure}
\end{figure}

Figure \ref{fig:res_exp1} shows the results of \textbf{Experiment 1}. Here we observe the average predicted probability and $\frac{1}{4}$ of one standard deviation across the three visits, shown in \textcolor{red}{red}, and the predicted probability after applying the inverse classification process at $v=1,2$: \textcolor{blue}{blue} shows evaluation of the probability using $\bx_{v_U}$ and \textcolor{magenta}{magenta} shows evaluation of the probability using $\bx_{v+1_U}$. We first observe that the inverse classification process was successful in reducing P(CVD) at both visits ($v=1,2$). Secondly, we observe that probabilistic results differ between \textcolor{blue}{blue} and \textcolor{magenta}{magenta}, which directly supports our hypothesis that accounting for recommendation implementation time leads to different probability estimates.

\begin{figure}[H]
	\centering
	\includegraphics[scale=.39]{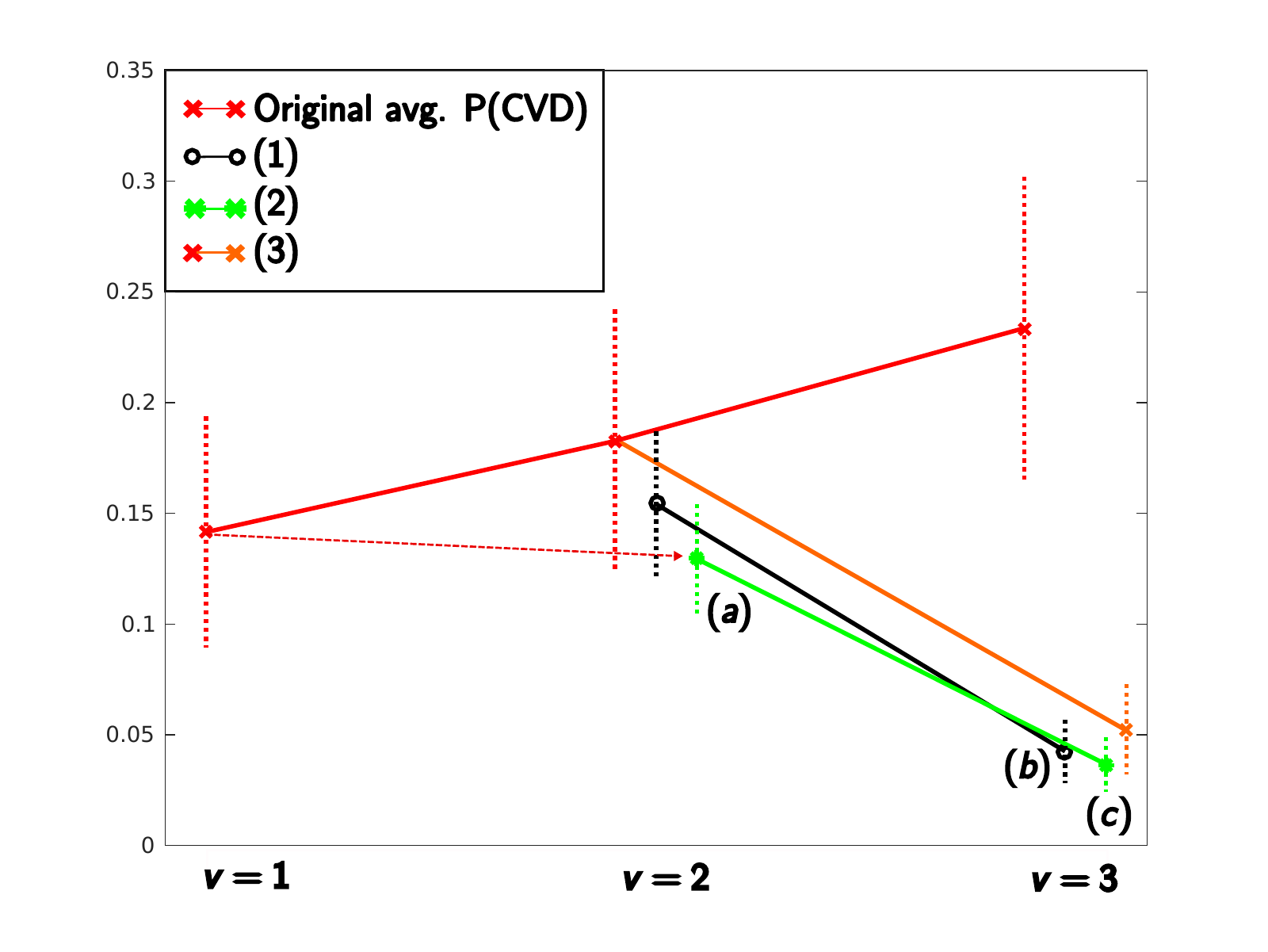}
	\caption{Results from Experiment 2. \label{fig:res_exp2}}
\end{figure}

Figure \ref{fig:res_exp2} shows the results of \textbf{Experiment 2}. Here, we present results in a similar manner to Experiment 1, but with some slight differences; the means and $\frac{1}{4}$ of one standard deviation are represented as they previously were, and \textcolor{red}{red} still denotes the original predicted probability. \textbf{(1)}, represented in black, starts with the predicted probability at $v=2$, taking into account $r_v=1$. The inverse classification process is then applied, leading to the predicted probability at $v=3$, represented by (b) -- this corresponds to (b) in Figure \ref{fig:experiments}. \textbf{(2)}, represented in \textcolor{green}{green}, begins with optimized instances from $v=1$, corresponding to (a) in Figure \ref{fig:experiments}, and ending with (c) in the same figure. \textbf{(3)}, shown in \textcolor{orange}{orange}, is the predicted probability at $v=3$ taking both $r_1$ and $r_2$ into account.

The first red, black, and orange data points in Figure \ref{fig:res_exp2} show the average predicted probability of CVD, taking past risk into account (i.e.,~they parallel what is shown in \textcolor{red}{red}, differing only in that past risk has been incorporated). Comparing the black point at $v=2$ and orange point at $v=3$ we see that there are marked differences between the unaccounted for risk predictions (red) and those that take past risk into account: past risk incorporation has lead to lower probability estimates. The lower estimates are intuitive, as our learning method has become more certain of those individuals who will not develop CVD -- low past risk likely indicates no immediate threat of developing a disease that takes years to manifest.  As Table \ref{tab:dset_sum} shows, $\approx 2$\% of instances develop CVD at each $v$, which supports the lower average probability estimates.

The same holds true for the two green and one black data points in Figure \ref{fig:res_exp2}. These points represent averages (and $\frac{1}{4}$ of one standard deviation $\pm$ the mean) obtained after applying inverse classification, accounting for past risk and using $\bx_{v+1_U}$ in the estimates. We observe only slight improvements in mean CVD probability after performing the process, versus the more extreme values observed in Figure \ref{fig:res_exp1}. Additionally, we can see that (b) and (c) in Figure \ref{fig:res_exp2} are very similar -- (c) is $.4\%$ lower on average.  This is likely the result of diminishing returns. Once appropriate lifestyle adjustments have been made, making further adjustments are unlikely to be as beneficial.

%\begin{figure}
%	\centering
%	\includegraphics[scale=.30]{figures/risk_opt_plot_new.pdf}
%	\caption{Results from Experiment 2. \label{fig:res_exp2}}
%\end{figure}

\section{Conclusions}

In this work we proposed two ways in which longitudinal data can be fused with an existing inverse classification framework to arrive at more realistic assessments of risk. First, we used past risk as an immutable predictive feature in the present. Second, after making recommendations, to reflect the fact that implementation and benefit are not observed instantaneously, we used the unchangeable features from the next discrete time period to estimate probability improvements. In our experiments we noticed that the inclusion of such factors resulted in different probability estimates than those obtained without. Future work in this area may benefit from the use of methods that are capable of taking all prior and current features into account, as well as future unchangeable features, which may cumulative help in making the inverse classification process as precise as possible.

\renewcommand{\bibname}{References}
\bibliography{lash_pakdd_2017}
\bibliographystyle{splncs03}

\end{document}